%% file: main.tex
\documentclass[10pt,twocolumn,letterpaper]{article}

\usepackage[pagenumbers]{cvpr} 
\usepackage{graphicx}
\usepackage{multirow}
\usepackage{caption}
\usepackage{cite}
\usepackage{bm}
\usepackage{amsmath}
\usepackage{amssymb}
\usepackage{longtable}
\usepackage{tabularx}
\usepackage{indentfirst}
\usepackage{makecell}
\usepackage{colortbl}
\usepackage{arydshln}
\usepackage{fontawesome5}
\definecolor{lightred}{rgb}{1.0, 0.8, 0.8}
\definecolor{lightorange}{rgb}{1.0, 0.9, 0.8}
\definecolor{lightyellow}{rgb}{1.0, 1.0, 0.8}
\newcommand{\ud}[1]{\underline{#1}}

\input{preamble}

\definecolor{cvprblue}{rgb}{0.21,0.49,0.74}
\usepackage[pagebackref,breaklinks,colorlinks,citecolor=cvprblue]{hyperref}

\def\paperID{*****} 
\def\confName{3DV\xspace}
\def\confYear{2025\xspace}

\title{OG-Mapping: Octree-based Structured 3D Gaussians for Online Dense Mapping }
\vspace{-5mm}
\author{Meng Wang\textsuperscript{1,2}, Junyi Wang\textsuperscript{3}, Changqun Xia\textsuperscript{2}, Chen Wang\textsuperscript{4}, Yue Qi\textsuperscript{1} \\
    \textsuperscript{1}State Key Laboratory of Virtual Reality Technology and Systems, Beihang University.\\
    \textsuperscript{2} PengCheng Laboratory. 
    \textsuperscript{3} School of Computer Science and Technology, Shandong University.\\
    \textsuperscript{4} Beijing Technology and Business University.
}

\begin{document}
\twocolumn[{
    \renewcommand\twocolumn[1][]{#1}%
    \maketitle
    \begin{center}
        \centering
        \vspace{-20pt}
        \includegraphics[width=\textwidth]{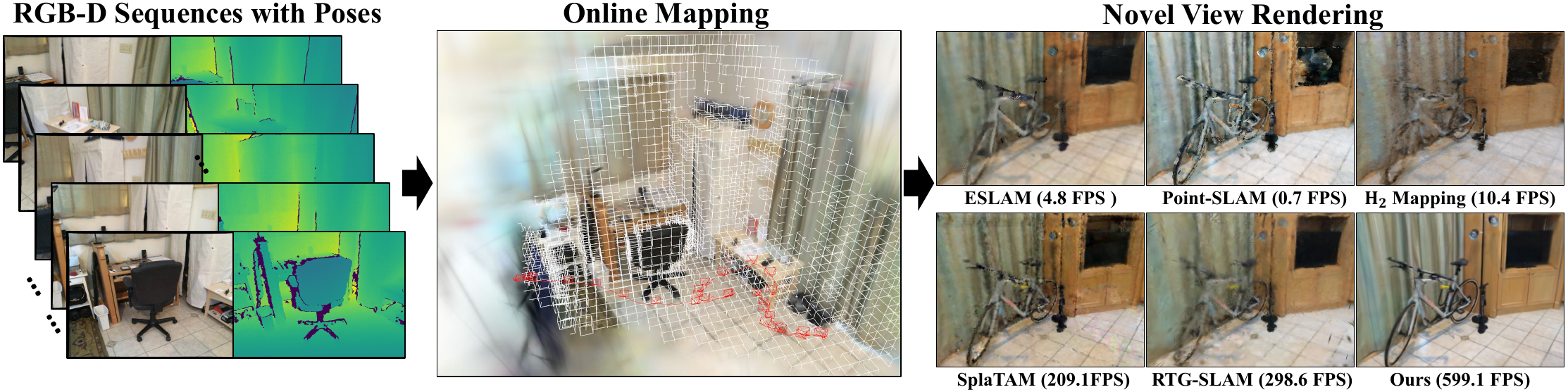}
        \vspace{-20pt}
        \captionof{figure}{In this work, we introduce OG-Mapping, a novel online dense mapping framework with an octree-based structured 3D Gaussian representation. 
        By combining our proposed anchor-based progressive map refinement strategy and dynamic keyframe window, OG-Mapping achieves fast, high-fidelity online reconstruction with efficient memory usage, and demonstrates superior realism in novel view synthesis compared to other existing RGB-D online mapping methods. The rendering FPS is indicated to the right of each method.
        }\vspace{-10pt}
        \label{fig:teaser}
    \end{center}
}]

\input{sec/0_abstract}    
\input{sec/1_intro}
\input{sec/2_related_work.tex}

\input{sec/3_method.tex}
\input{sec/4_experiment.tex}

\input{sec/5_conclusion.tex}
{
    \small
    \bibliographystyle{ieeenat_fullname}
    \bibliography{main}
}

\end{document}

%% file: preamble.tex
\usepackage[dvipsnames]{xcolor}


%% file: sec/0_abstract.tex
\begin{abstract}
\vspace{-10pt}
3D Gaussian splatting (3DGS) has recently demonstrated promising advancements in RGB-D online dense mapping.
Nevertheless, existing methods excessively rely on per-pixel depth cues to perform map densification, which leads to significant redundancy and increased sensitivity to depth noise. Additionally, explicitly storing 3D Gaussian parameters of room-scale scene poses a significant storage challenge.
In this paper, we introduce OG-Mapping, which leverages the robust scene structural representation capability of sparse octrees, combined with structured 3D Gaussian representations, to achieve efficient and robust online dense mapping.
Moreover, OG-Mapping employs an anchor-based progressive map refinement strategy to recover the scene structures at multiple levels of detail. Instead of maintaining a small number of active keyframes with a fixed keyframe window as previous approaches do, a dynamic keyframe window is employed to allow OG-Mapping to better tackle false local minima and forgetting issues. 
Experimental results demonstrate that OG-Mapping delivers more robust and superior realism mapping results than existing Gaussian-based RGB-D online mapping methods with a compact model, and no additional post-processing is required.
\end{abstract}

%% file: sec/1_intro.tex
\vspace{-20pt}
\section{Introduction}
\label{sec:intro}
\vspace{-5pt}
Constructing highly detailed dense maps in real-time holds significant importance in AR/VR, robotics, and digital twins applications.  For the past several decades, research on mapping has extensively centered around the scene representation, resulting in various representations such as occupancy grids \cite{octomap}, TSDF \cite{scalable, online}, surfels \cite{elasticfusion, real} and point clouds \cite{interactive}. 
Although systems utilizing these map representations have reached a production-ready standard over the past years, there remain notable deficiencies that demand attention and resolution.
These methods exhibit deficiencies in rendering realism of reconstructed results, and fine-detailed maps based on the above representations require massive amounts of storage space.


In recent years, implicit representations \cite{neus, ngp, triplane} have demonstrated promising outcomes in various domains following the advent of Neural Radiance Fields (NeRF)  \cite{nerf}.
Numerous studies \cite{imap, niceslam, h2mapping, coslam, eslam, plgslam} employ these implicit representations to enhance mapping methodologies and exhibit strengths in generating high-quality dense maps with low memory consumption.
Nevertheless, volumetric sampling significantly limits these methods' efficiency. Consequently, they opt for optimization over a sparse set of pixels instead of dense per-pixel photometric error, resulting in the reconstruction dense maps lack the richness and intricacy of texture details.

More recently, several works \cite{scaffoldgs, mipgs} based on the 3D Gaussian representation \cite{3dgs} and tile-based splatting technique have shown great superiority in the efficiency of differentiable rendering. 
However, these methods are meticulously designed for offline reconstruction scenarios. 
Concurrent works \cite{splatam, monogs, yugay2023gaussianslam, rtg, photo, yan2023gs} attempt to incorporate 3D Gaussians representation into RGB-D online dense mapping to deliver high-fidelity rendering, overcoming the limitation of Nerf-based methods. While promising results are demonstrated, these methods neglect the scene structure and rely excessively on pixel-level depth information to expand 3D Gaussians, resulting in significant redundancy and being sensitive to depth noise effects.

To overcome these challenges, in this paper, we introduce an innovative framework named OG-Mapping to perform efficient and highly detailed online dense mapping.
Our solution comprises three main components. First, we incrementally build a sparse voxel octree with Morton coding  \cite{vespa2018efficient} for fast allocation and retrieval of anchors to dynamically expand the map.  
Each anchor tethers a set of 3D Gaussians with learnable offsets. 
The attributes (color, opacity, quaternion, and scale) of these Gaussians are predicted based on the viewing direction and anchor feature encodings.  
By leveraging this structured representation, we can effectively mitigate pixel-level depth noise effects and avoid the substantial memory consumption associated with explicitly storing large amounts of 3D Gaussian attributes.
Second, as the anchors organized by the sparse octree can only provide a rough description of the scene structure, we further design an anchor-based progressive map refinement strategy to recover the scene structures at different levels of detail by adaptively adding finer-level anchors.
Unlike the previous method \cite{scaffoldgs} utilizes an error-based policy, when an area is marked as under-optimized, we grow anchors according to the level hierarchy of anchors already assigned to that area, resulting in a more compact map.
Finally, we develop a dynamic keyframe window to alleviate the false local minima and forgetting issue to improve mapping quality.
Through extensive experiments, we empirically demonstrate that our method achieves superior mapping results with approximately a 5dB enhancement in PSNR on real-world scenes \cite{scannet, rtg}, with a more compact model, while maintaining a more compact model and operating at high processing speed.

To summarize, our contributions are as follows:
\begin{itemize}
  \item We introduce a novel framework that naturally integrates the sparse octree and structured 3D Gaussian representation to perform efficient and detailed online dense mapping.
  \item We design an anchor-based progressive map refinement strategy for better scene coverage and mapping quality.
  \item We develop a dynamic keyframe window to mitigate the false local minima and forgetting issue.
\end{itemize}

%% file: sec/2_related_work.tex
\section{Related Work}
\label{sec:related_work}
\noindent
\textbf{Classical RGB-D online dense mapping.}
For online 3D reconstruction of scenes, 
various explicit representations have been employed to store scene information,
 including point clouds \cite{interactive}, surface elements(surfels) \cite{elasticfusion, real} and truncated signed distance functions(TSDF) \cite{scalable, online}. 
 Several works \cite{tandem,cnn} leverage deep learning to improve the accuracy and robustness
 of above mentioned representations, and even achieve dense reconstruction with monocular input \cite{droid,d3vo}.
 These methods are renowned for their rapid processing speed, which is attributable to the inherent 
physical properties of explicit representations.
However, they demand significant memory resources to manage high-detail mapping \cite{shine},
and are incapable of realistically rendering from novel views.  

\begin{figure*}[th]
    \begin{center}
    \includegraphics[width=0.83\linewidth]{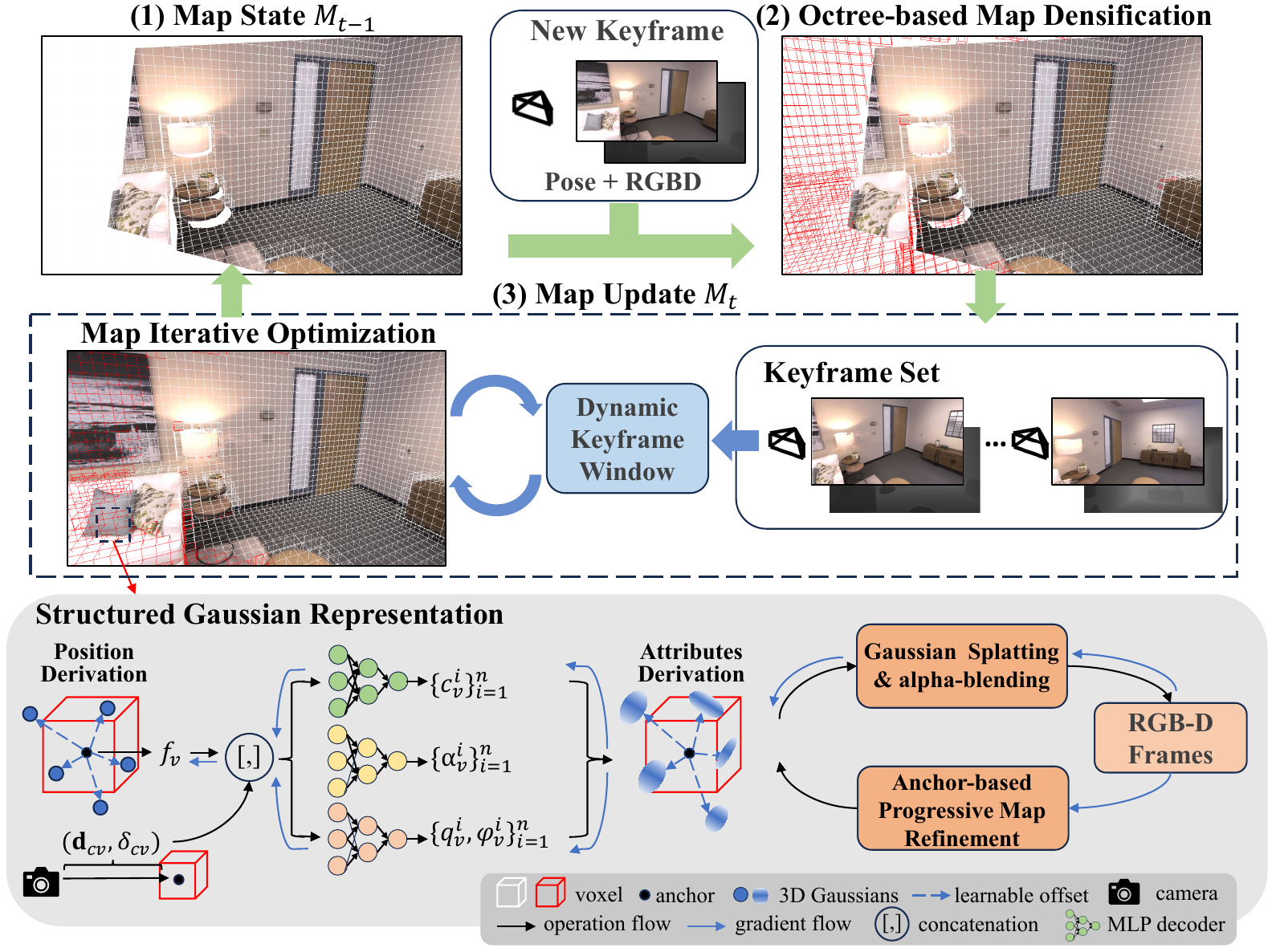}
    \end{center}
    \vspace{-15pt}
       \caption{ \textbf{Overview of OG-Mapping.} Given a set of sequential RGB-D frames and camera poses, we utilize an octree-based structured 3D Gaussians as the scene representation to perform efficient online dense mapping. 
       When a new keyframe is detected, we employ a sparse octree to swiftly capture the rough structure of the new observed region to guide anchor densification (Sec. \ref{sec: represention}) . 
       During the map update process, we perform anchor-based progressive map refinement to enhance the geometry and appearance quality (Sec. \ref{sec: optimization}), 
       and construct a dynamic keyframe window to effectively mitigate false local minima and forgetting issues (Sec. \ref{sec: keyframe}).
       }
    \label{fig: overview}
    \vspace{-15pt}
\end{figure*}

\noindent
\textbf{NeRF-based RGB-D online dense mapping.}
Following the significant success of neural radiance fields (NeRF)  \cite{nerf}, 
several studies leveraged latent features and neural networks as implicit representations to integrated NeRF with RGB-D dense mapping.
iMAP  \cite{imap} presents the first NeRF-style online dense mapping, using Multi-Layer Perceptrons (MLP) as the scene representation. 
NICE-SLAM  \cite{niceslam} represents scenes as hierarchical feature
grids, utilizing pre-trained MLPs for decoding. 
To enhance mapping speed and expand representational capacity, 
various representations have been investigated, including multi-resolution hash grids \cite{h2mapping, coslam, h3mapping}, factored grids \cite{eslam}, 
sparse octree grids \cite{voxfusion} and neural point clouds \cite{pointslam}.
Nevertheless, the aforementioned methods struggle to achieve fine-detailed rendering results and maintain fast rendering speed.
These limitations arise from their reliance on time-consuming volumetric rendering techniques.
In contrast, our approach leverages fast rasterization, enabling complete use of
per-pixel dense photometric errors. 

\noindent
\textbf{Gaussian-based RGB-D online dense mapping.}
The high-fidelity and rapid rasterization capabilities of 3D Gaussian Splatting (3DGS) \cite{3dgs} facilitate superior quality and efficiency in scene reconstruction.
Recently, many works \cite{splatam,monogs,photo, yugay2023gaussianslam} have attempted to apply 3DGS in online dense mapping. 
SplaTAM  \cite{splatam} adopts an explicit volumetric approach using isotropic Gaussians, enabling
precise map densification. However, this method necessitates projecting and densifying each pixel in the depth image to perform gaussian densification, resulting in substantial map storage usage.
MonoGS \cite{monogs} randomly select a subset of pixels for projection, which necessitates more optimization time for map densification. 
Concurrent to our work,  
CG-SLAM  \cite{cg} introduces a depth uncertainty model to select more valuable Gaussian primitives during optimization.
RTG-SLAM \cite{rtg} treats each opaque Gaussian as an ellipsoid disc
 on the dominant plane of Gaussian to maintain a compact representation. 
 NGM-SLAM \cite{ngm} and Gaussian-SLAM \cite{yugay2023gaussianslam} focuse on building submap for 3D Gaussian represention. 
 More recently MG-SLAM \cite{mgslam} leverages Manhattan World hypothesis and additional semantic informations to refine and complete scene geometries.
Different from these methods, our approach is based on the structured 3D Gaussian representation \cite{scaffoldgs} and utilizes the structured information of the octree to guide the distribution of Gaussian kernels, resulting in better robustness to depth noise and smaller map size.

%% file: sec/3_method.tex
\vspace{-10pt}
\section{Method}
\vspace{-5pt}
The overview of OG-Mapping is shown in Fig. \ref{fig: overview}. 
Taking RGB-D images from sensors and poses from other tracking modules, we utilize a structured 3D Gaussians representation \cite{scaffoldgs} managed by a sparse octree(Sec. \ref{sec: represention}) to depict the scene geometry and appearance. 
During mapping process, we employ an anchor-based progressive map refinement strategy to enhance map reconstruction quality(Sec. \ref{sec: optimization}). 
In Sec. \ref{sec: keyframe}, we introduce how to build our dynamic keyframe window to mitigate the fasle local minima and forgetting problem. Sec. \ref{sec: losses} elaborate the online optimization details.
\subsection{Structured Gaussian Representation for Online Mapping}
\label{sec: represention}
\noindent
\textbf{Scene Representation.}
We represent the underlying map of the scene as a set of anchors \cite{scaffoldgs}. Specifically, we first voxelize the scene using the provided camera pose and depth image. For each voxel, the center $v$ is treated as an anchor point, equipped with a level mark $l_v \in \mathbb{N}_0$, a scaling factor $s_v\in \mathbb{R}^{3}$, 
$n$ learnable offsets $\{ o_{v}^i \}^{n}_{i=1} \in \mathbb{R}^{n \times  3}$ and a feature vector $f_v = \textbf{encoding}(v)$ 
(In our dense version, the encoding function is the multi-hash encoding \cite{ngp}, while in the sparse version, it represents local context encoding. Implementation details are in supplementary). 

For each visible anchor within the viewing frustum, $n$ 3D Gaussians are generated. 
The positions ${\{\mu_{v}^i \}}^{n}_{i=1}$ of these 3D Gaussians are calculated as:
\begin{equation}
    {\{\mu_{v}^i \}}^{n}_{i=1} = p_{v} + s_v  \cdot {\{o_{v}^{i} \}}_{i=1}^{n}
\end{equation}
The other attributes (color $c_v^i\in \mathbb{R}^3$, opacity $\alpha _v^i \in \mathbb{R_{>0}}$, quaternions $q _v^i\in \mathbb{R}^4$, and scale $\varphi _v^i\in \mathbb{R}^3$) of $n$ 3D Gaussians are decode from the relative distance $\delta _{cv}$, view direction $\textbf{d} _{cv}$ , and the anchor feature $f_v$ using individual multilayer perceptron(MLP) decoders, denoted as $F_{color}$, $F_{opacity}$ and $F_{cov}$:

\begin{equation}
    \begin{array}{c}
        {\{ c_v^i \}}_{i=1}^{n} = \textbf{F}_{color}(\delta _{cv},\textbf{d}_{cv},f_v), \\
        {\{ \alpha_v^i \}}_{i=1}^{n} = \textbf{F}_{opacity}(\delta _{cv},\textbf{d}_{cv},f_v), \\
        {\{ q _v^i, \varphi  _v^i\}}_{i=1}^{n} = \textbf{F}_{cov}(\delta _{cv},\textbf{d}_{cv},f_v), \\
    \end{array}
\end{equation}
where the relative distance $\delta _{cv}$ and the viewing direction $\textbf{d}_{cv}$ between camera position $p_{c}$ and anchor point position $p_{v}$ are calculated as follows :
\begin{equation}
    \delta _{cv} = {\Vert p_{c} - p_{v} \Vert}_{2}, \quad \textbf{d}_{cv} =  \cfrac{p_{c} - p_{v}}{\delta _{cv}}.
\end{equation}
Then, these generated $\mathcal{N}$ 3D Gaussians are used for fast rasterization rendering to produce color and depth maps. Given a viewpoint, the rendered color of each pixel $\textbf{p}$ can be written as:
\begin{equation}
    C(\textbf{p}) = \sum _{i\in \mathcal{N}}{c^i\sigma ^i\prod_{j=1}^{i-1} (1-\sigma ^j)}, \sigma^i = \alpha^iG_{2d}^i(\textbf{p}),
\end{equation}
where the 2D Gaussians $G_{2d}(\textbf{p})$ are transformed from 3D Gaussian $G(\textbf{p})$ introduced by  \cite{3dgs}.
Similarly, per-pixel depth is rendered via alpha-blending:
\begin{equation}
    D(\textbf{p}) = \sum _{i\in \mathcal{N}}{z^i\sigma ^i\prod_{j=1}^{i-1} (1-\sigma ^j)},
\end{equation}
where $z^i$ is the distance of the $i^{th}$ 3D Gaussian's position $u^i$ along the camera ray.

\noindent
\textbf{Map Densification.}
To achieve a comprehensive representation of the environment, we need to add new anchors to the scene during online scanning to cover newly observed regions. 
The adaptive map densification approach in  \cite{splatam, rtg}, which is based on pixel-level projection error, is sensitive to depth noise and is unreliable when handling edges.
Upon receiving a new keyframe, we dynamically allocate new voxels using the provided pose and depth image, 
incrementally updating a sparse octree to roughly encompass all visible regions.

\subsection{Anchor-based Progressive Map Refinement}
A progressive optimization strategy can better shape the loss landscape, reducing the risk of the algorithm becoming trapped in misleading local minima. 
Such a strategy has been successfully applied in various computer vision applications like registration \cite{barf} and surface reconstruction \cite{neuralangelo}. 
OG-Mapping also utilizes a coarse-to-fine optimization scheme to reconstruct the scene with progressive levels of detail, by adaptively adding finer-level anchors to under-optimized regions.

\label{sec: optimization}
\begin{figure}[thp]
    \begin{center}
    \includegraphics[width=0.98\linewidth]{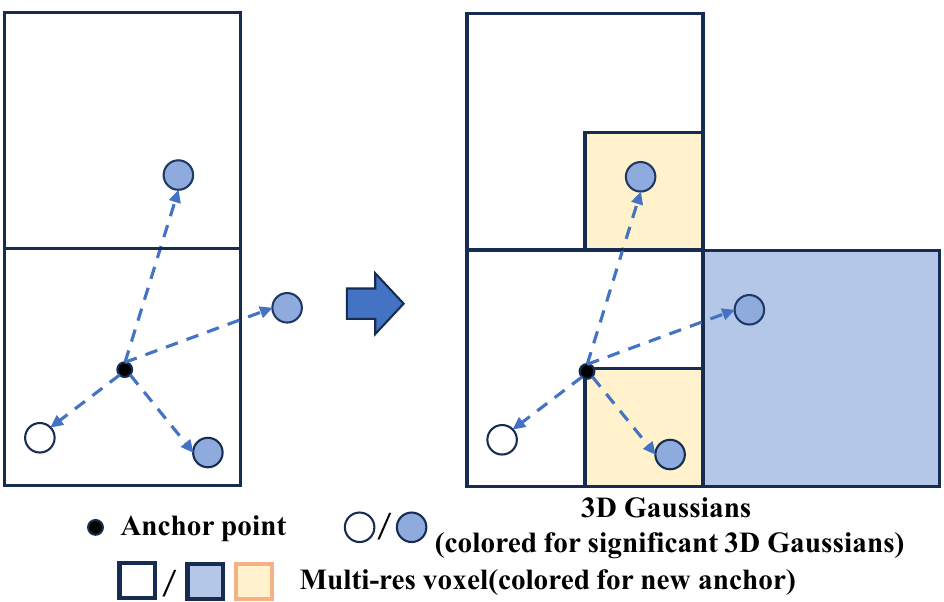}
    \end{center}
    \vspace{-15pt}
    \caption{\textbf{Anchor-based Progressive Map Refinement.} We grow new anchors in under-optimized regions based on the level of 3D Gaussians and their gradients. If the regions already contain anchors at the same level, the granularity of these new anchors will be increased.}
    \vspace{-15pt}
\label{fig:anchor_growing}
\end{figure}
To ensure system efficiency, the sparse octree only maintains the coarsest granularity of anchors to quickly reveal scene structure information. 
The 3D Gaussians inferred from these newly added anchors have a large scale, which makes it difficult to fit areas with high-frequency texture changes. Therefore, anchors with finer granularity need to be added to these under-optimized areas.
Scaffold-GS \cite{scaffoldgs} employs an error-based anchor growth method, which allows the 3D Gaussians of fine-level anchors to participate in the growth of coarser ones. Nevertheless, during online reconstruction, it fails to provide sufficient optimization iterations to stabilize the gradients of such fine-level 3D Gaussians, resulting in the generation of numerous unnecessary anchors.

To address the aforementioned issues, we develop an anchor-based progressive map refinement strategy as illustrated in  Fig.  \ref{fig:anchor_growing}. Specifically,
for each level $l$, we predefine the corresponding voxel size $\gamma _{\,l}$ and significance threshold $\tau_{\,l}$.
To determine under-optimized areas, the gradients of 3D Gaussians belonging to anchor $v$ are denoted as $\{\bigtriangledown_{v}^{i} \}_{i=1}^n$. 
Subsequently, the 3D Gaussian satisfying $\bm{\bigtriangledown}_{v}^{i} \geqslant  \tau _{l_v}$ marks its respective region as candidate region.
A new anchor $v'$ with level $l_{v'} = l_v$ is deployed at the position $u_v^i$. 
If the candidate area already has anchor with level $l_{v'}$, indicating that the current level of detail is still insufficient to meet the optimization requirements, the granularity of anchor $v'$ will be increased.
Specifically, the 3D Gaussians that have already participated in growth will not be used for further anchor growth.
In practice, the variations of voxel size and significant threshold between different levels are defined as:
\begin{equation}
    \gamma _{\,l+1} = \gamma _{\,l} \, / \,4 ,\quad \tau_{\,l+1} = \tau_{\,l} * 2.
\end{equation}
This coarse-to-fine strategy allows us to effectively suppress the expansion of anchors caused by gradient instability.
\subsection{Dynamic Keyframe Window Construction}
\label{sec: keyframe}

\noindent
\textbf{Keyframe Insert Selection.} The keyframe set is constructed based on the overlap between the visible coarsest anchors of the current frame and the last keyframe.
A new input RGB-D frame is added to the keyframe set if the ratio $N_{overlap} / N_{union}$ smaller than a threshold, where $N_{overlap}$ is the number of coarsest granularity anchors visible in both the current frame and the last keyframe, and $N_{union}$ represents  the total number of visible coarsest anchors of them.
By this insertion strategy, we ensure the views in the keyframe set have relatively little overlap.

\noindent
\textbf{Dynamic Keyframe Window.}
When a new keyframe is added, it typically indicates that new unexplored areas require optimization. 
Relying solely on the data from this single keyframe for map optimization can lead to severe forgetting and overfitting issues, 
resulting in poor final reconstruction quality. Existing online mapping methods usually select a specific number of keyframes from the keyframe set, along with the newly added keyframe(some works additionally add the most recent keyframe), to form a fixed keyframe window. Only the keyframes within this fixed window are used for map optimization.
\begin{figure}[thp]
    \begin{center}
    \includegraphics[width=1.0\linewidth]{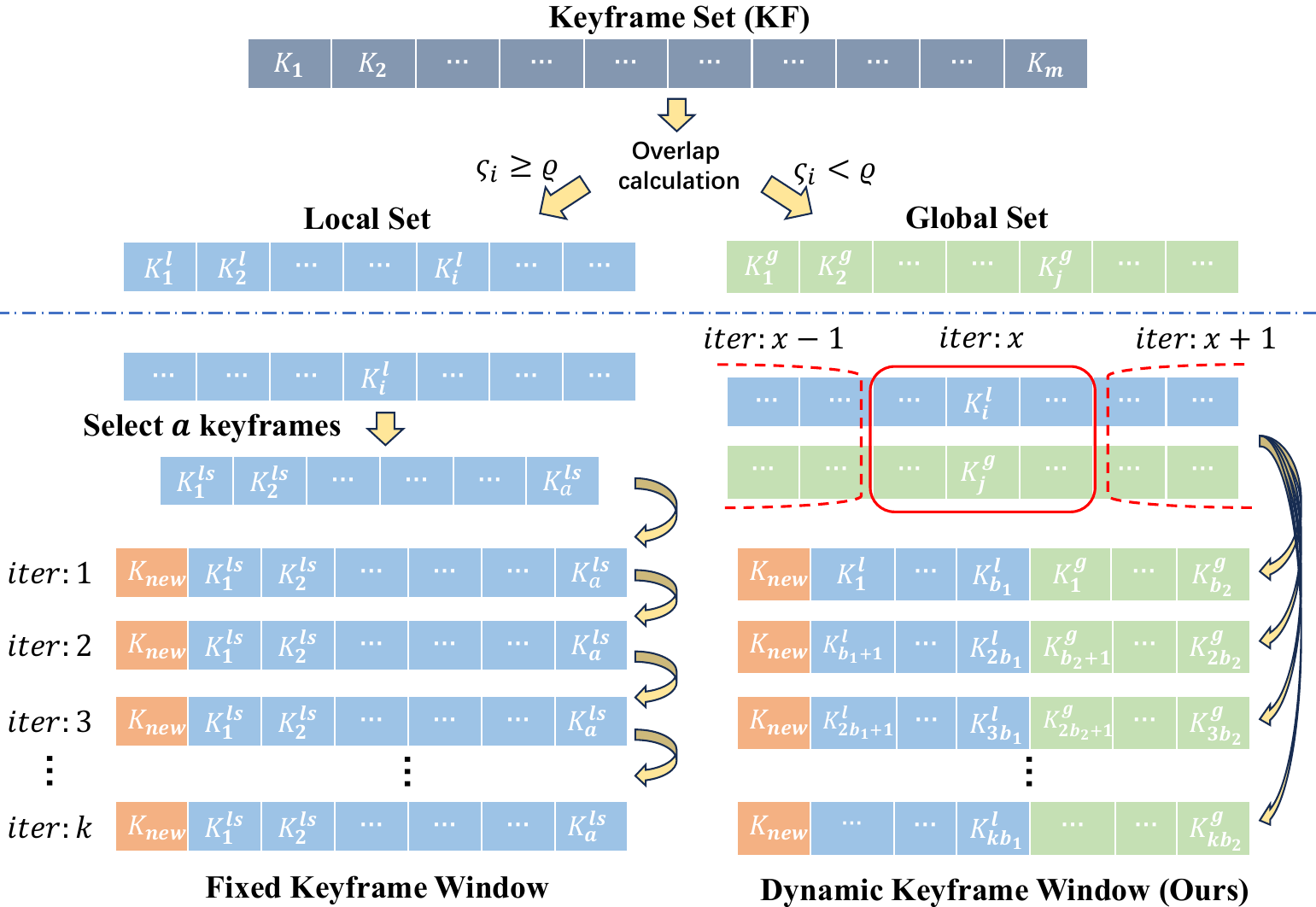}
    \end{center}
    \vspace{-15pt}
    \caption{\textbf{Illustration of the differences between Fixed Keyframe Window and our Dynamic Keyframe Window.} Left: Fixed Keyframe Window maintains a static window content across all optimization iterations. Right: our dynamic Keyframe Window updates the keyframe window with new data each optimization iteration.
       }
    \vspace{-10pt}
\label{fig:keyframe_select}
\end{figure}
However, these methods face challenges in balance optimization iterations with mapping quality. Suppose the new keyframe $K_{new}$ observes a new region with distribution $\zeta$.
Since the information for $\zeta$ is only observed by $K_{new}$, 
it is necessary to optimize $K_{new}$ with $k$ times to accurately fit the new region. Let $a$ represent the number of keyframes in current window, the total number of inference operations required is $k \times (a+1)$. 
A diminutive $b$ is prone to causing significant overfitting and forgetting problem, which hampers the model's performance on previously visited area. Conversely, optimize $K_{new}$ with a larger $b$ can lead to an excessively high number of inference operations, potentially causing the model to become computationally intensive and time-consuming.

Further considering this issue, it becomes apparent that only the $K_{new}$ necessitates multiple optimization iterations. 
The role of the other keyframes is to alleviate local overfitting and forgetting issues.
Thus, having diverse sources for these keyframes is advantageous. 
Based on the above analysis, instead of using a fixed keyframe window, we develop a simple yet effective dynamic keyframe window.
Specifically, we first calculate the overlap weight $\varsigma $ between the existing keyframes in the keyframe set $\textbf{KF}$ and the $K_{new}$. $\textbf{KF}$ is subsequently divided into two sets with threshold $\varrho$: the local set $(\{K_{i}|K_{i} \in \textbf{KF}, \varsigma _{i} \geq \varrho \})$ and global set $(\{K_{i}|K_{i} \in \textbf{KF}, \varsigma _{i} < \varrho \})$.
During each new optimization iteration, we clear the keyframe window of all keyframes except $K_{new}$. 
Then, we select $b_1$ keyframes from the local set and $b_2$ keyframes from the global set without replacement and add them to the keyframe window ($a=b_1+b_2$). 
By employing the aforementioned method, we ensure that the contents of the keyframe window are always dynamically changing. 
This approach enables effective adaptation to newly observed regions while incorporating past experiences, thereby addressing the limitations of the fixed keyframe window method.
A more intuitive comparison between the two methods can be found in Fig. \ref{fig:keyframe_select}.

\subsection{Losses Design}
\label{sec: losses}
The learnable parameters and MLPs are optimized with respect to the $L1$ loss over rendered pixel colors , denoted as $\mathcal{L}_{c}$, and depths, denoted as $\mathcal{L}_{d}$. The loss function is further augmented by SSIM term \cite{ssim} $\mathcal{L}_{SSIM}$ and scale term \cite{mixture} $\mathcal{L}_{s}$:
\begin{equation}
    \mathcal{L} = \lambda_{c}\mathcal{L}_{c} + \lambda_{SSIM}\mathcal{L}_{SSIM} + \lambda_{d}\mathcal{L}_{d} + \lambda_{s}\mathcal{L}_{s}.
\end{equation}
where the Gaussians scale term $\mathcal{L}_s$ is:
\begin{equation}
    \mathcal{L}_{s} = \sum_{i = 1}^{\mathcal{N}} Prod(\varphi^i) .
\end{equation}
The $Prod(.)$ is the product computation function and $\mathcal{N}$ is the the number of all 3D Gaussians.

%% file: sec/4_experiment.tex
\section{Experiment}

\subsection{Experimental Setup}{
\noindent
\textbf{Dataset.}{
To evaluate the performance of our proposed method, we compare its mapping accuracy and time consumption with
other RGB-D mapping systems currently open-source on both the synthetic Replica dataset  \cite{replica} and the real-world ScanNet dataset
 \cite{scannet} in addition to a self-scanned scene provided by  \cite{rtg}.
}

\noindent
\textbf{Metrics.}{
Following the evaluation protocol of mapping results used in SplaTAM \cite{splatam}, for measuring RGB rendering performance, we report standard photometric rendering quality metrics:PSNR \cite{psnr}, SSIM \cite{ssim} and LPIPS(AlexNet) \cite{lpips}. Depth rendering performance is measured by Depth L1 loss(cm). All rendering metrics are computed on each frame with valid pose to evaluate the map quality. We report the average across five runs for
all our evaluations.
}

\noindent
\textbf{Baseline Methods.}{
  We select several advanced Gaussian-based dense RGB-D SLAM methods currently open-source, MonoGS \cite{monogs}, SplaTAM \cite{splatam} and RTG-SLAM \cite{rtg} for comparison.
  We also benchmark our method against other RGB-D based approaches  \cite{h2mapping, eslam, pointslam, h3mapping} that, like ours, do not have explicit loop closure and submap.
}

\noindent
\textbf{Implementation Details.}{
  All methods are benchmarked on a desktop computer with an intel i7-14700K CPU and an Nvidia RTX 4090 GPU.
  As we exclusively focus on incremental mapping, we omit the tracking component of  \cite{eslam, pointslam, splatam, monogs} and instead employ the ground truth pose. All methods are evaluated using their single-threaded versions.
  Specifically, the scene refinement module used after online mapping process in MonoGS \cite{monogs} was removed to ensure a fair comparison. An anchor is pruned if all the 
  opacity of its 3D Gaussians is less than $\rho =0.01$. 
  To provide a more comprehensive comparison with other methods, in addition to the full version(marked as 'Ours'), we also offer the 'Ours-sparse' version (using a larger 
$\rho$ to prune more anchors) and the 'Ours-sparse\faRunning' version (based on the last version but using only half the number of iterations for mapping).
  More details of hyperparameters are provided in the supplementary material. 
}

}

\begin{table}[h]
  \renewcommand{\arraystretch}{1.0}
  \setlength{\tabcolsep}{3pt}
  
  \resizebox{\linewidth}{!}{
  \vspace{-10pt}
      \begin{tabular}{c|ccccccccccc}
          \hline
          \multicolumn{1}{c}{\textbf{Methods}}  &  \multicolumn{1}{l}{Metrics}  & {R0} & {R1} & {R2} & {O0} & {O1} & {O2} & {O3} & {O4} & {Avg} \\
          \hline
          
          \multirow{3}{*}{\makecell{H$_{2}$-Mapping \cite{h2mapping}\\(RAL23)}} & 
                                  PSNR $\uparrow$           & 29.67                      & 32.34                      & 32.23                      & 37.87                        & 38.93                       & 31.02                        & 30.64                        & 33.07                         & 33.22                    \\
                                & SSIM $\uparrow$                   & 0.927                      & 0.956                      & 0.963                      & 0.982                & 0.984                       & 0.958                        & 0.95                         & 0.968                         & 0.932                    \\
                                & LPIPS $\downarrow$                 & 0.217                      & 0.168                      & 0.151                      & 0.09                        & 0.0945              & 0.203                        & 0.221                        & 0.166                         & 0.164                    \\
                                
          \hline
          \multirow{3}{*}{\makecell{ESLAM \cite{eslam}\\(CVPR23)}} & 
                                  PSNR $\uparrow$            & 26.40                      & 28.34                      & 30.25                      & 35.10                        & 34.76                        & 29.07                        & 28.83                        & 31.15                        & 30.49                    \\
                                & SSIM $\uparrow$            & 0.763                      & 0.874                    & 0.923                     & 0.928                        & 0.921                        & 0.879                        & 0.876                        & 0.908                        & 0.871                    \\
                                & LPIPS $\downarrow$         & 0.300                      & 0.292                     & 0.232                      & 0.180                        & 0.205                        & 0.235                        & 0.191                        & 0.199                        & 0.230                    \\

          \hline
          \multirow{3}{*}{\makecell{Point-SLAM \cite{pointslam}\\(ICCV23)}} & 
                                  PSNR $\uparrow$            & \ud{34.38}                      & \ud{35.05}                      & \ud{36.80}                      & 39.35                        & 40.29                        & \ud{34.95}                        & \ud{34.54}                        & 35.66                        & \ud{36.38}                    \\
                                & SSIM $\uparrow$                 & 0.937                 & 0.942                      & 0.955                     & 0.962                        & 0.960                        & 0.916                        & 0.917                        & 0.942                        & 0.941                    \\
                                & LPIPS $\downarrow$              & 0.093                 & 0.106                      & 0.102                     & 0.088                        & 0.103                        & 0.147                        & 0.116                        & 0.129                        & 0.111                    \\

          \hline
          \multirow{3}{*}{\makecell{H$_{3}$-Mapping \cite{h3mapping}\\(arXiv24)}} & 
                                  PSNR $\uparrow$            & 33.16                      & 34.99                      & 35.24                      & 39.85                        & 40.12                        & 33.89                        & 34.10                        & 35.99                        & 35.92                    \\
                                & SSIM $\uparrow$                 & 0.921                 & 0.939                      & 0.948                     & 0.969                        & 0.966                        & 0.936                        & 0.933                        & 0.950                        & 0.945                    \\
                                & LPIPS $\downarrow$              & -                 & -                      & -                     & -                        & -                        & -                        & -                        & -                        & -                    \\
          
          \hdashline
          
          \multirow{3}{*}{\makecell{MonoGS \cite{monogs}\\(CVPR24)}} 
                              & PSNR $\uparrow$               & 33.16                      & 34.99                      & 35.24                      & 39.85                        & 40.12                        & 34.59                        & 34.32                        & \ud{36.50}                        & 35.68                    \\
                                & SSIM $\uparrow$                  & 0.928                      & 0.928                      & 0.942                      & 0.972                        & 0.968                        & 0.941                        & 0.939                        & 0.952                  & 0.946                   \\
                                & LPIPS $\downarrow$                 & 0.128                      & 0.153                      & 0.129                      & 0.085                        & 0.084                       & 0.129                        & 0.107                        & 0.129                        & 0.118                    \\
          
          \hline
          \multirow{3}{*}{\makecell{SplaTAM \cite{splatam}\\(CVPR24)}} 
                              & PSNR $\uparrow$                     & 33.57                      & 34.40                      & 36.36                                & \ud{40.37}                        & \ud{40.51}                        & 33.38                        & 32.56                        & 35.07                        & 35.78                    \\
                                & SSIM $\uparrow$                  & \textbf{0.978}              & 0.973                      & \textbf{0.984}                      & 0.985                   & 0.980                        & \ud{0.974}                        & 0.968                        & 0.970                        & \ud{0.976}                   \\
                                & LPIPS $\downarrow$               & \ud{0.052}                       & \ud{0.064}                      & \textbf{0.053}                               & \ud{0.049}                  & \ud{0.067}                        & \ud{0.073}                        & \ud{0.081}                        & \ud{0.106}                        & \ud{0.068}                    \\
          \hline
          \multirow{3}{*}{\makecell{RTG-SLAM \cite{rtg}\\(SIGGRAPH24)}} 
                              & PSNR $\uparrow$               & 31.30                     & 33.95                      & 34.95                                       & 39.28                        & 39.63                        & 32.90                        & 32.86              & 36.12                        & 35.12                    \\
                                & SSIM $\uparrow$                  & 0.967                      & \textbf{0.979}                      & \ud{0.983}                      & \textbf{0.988}                   & \textbf{0.990}            & \textbf{0.981}              & \textbf{0.982}      & \textbf{0.985}               & \textbf{0.982}                   \\
                                & LPIPS $\downarrow$               & 0.144                       & 0.111                      & 0.116                               & 0.080                  & 0.097                        & 0.140                        & 0.133                    & 0.118                        & 0.117                    \\
          \hline
          \multirow{3}{*}{Ours} & PSNR $\uparrow$               & \textbf{36.24}          & \textbf{38.13}                      & \textbf{38.73}                      & \textbf{42.07}               & \textbf{42.26}                       & \textbf{36.15}                        & \textbf{36.34}               & \textbf{38.58}                        & \textbf{38.56}                    \\
                                & SSIM $\uparrow$                  & \ud{0.971}                   & \ud{0.974}                            & 0.976                               & \ud{0.986}                        & \ud{0.981}                                & 0.973                                 & \ud{0.971}                        & \ud{0.975}                        & \ud{0.976}                   \\
                                & LPIPS $\downarrow$                 & \textbf{0.044}         & \textbf{0.049}                  & \ud{0.055}                      & \textbf{0.036}               & \textbf{0.049}                       & \textbf{0.054}                        & \textbf{0.047}               & \textbf{0.050}                         & \textbf{0.048}                    \\
          \hline
          \end{tabular}
  }
  \vspace{-5pt}
  \caption{Quantitative comparison of our method against baselines for rendering results on the Replica \cite{replica} dataset. }
  \label{tab:replica_per}
  \vspace{-10pt}
\end{table}

\begin{table}[h]
  \renewcommand{\arraystretch}{1.0}
  \setlength{\tabcolsep}{3pt}
  \resizebox{\linewidth}{!}{
  \vspace{-10pt}
      \begin{tabular}{c|ccccccccc}
          \hline
          \textbf{Methods}   &  Metrics                     & 0000                & 0059                    & 0106                    & 0169                   & 0181                    & 0207                   & Avg \\
          \hline
          \multirow{3}{*}{\makecell{H$_{2}$-Mapping \cite{h2mapping}\\(RAL23)}} 
                                & PSNR $\uparrow$            & 21.02                      & 17.60                     & 15.59                      & 20.65                        & 18.83                       & 21.04                        & 19.12                                            \\
                                 & SSIM $\uparrow$            & \textbf{0.809}                      & 0.765                      & 0.742            & \ud{0.819}                        & \ud{0.844}                       & \textbf{0.822}                        & \ud{0.800}                                             \\
                                 & LPIPS $\downarrow$         & 0.475                      & 0.445                      & 0.522                      & 0.419                        & 0.492                       & 0.472                        & 0.471                                           \\
                                
          \hline
          \multirow{3}{*}{\makecell{ESLAM \cite{eslam}\\(CVPR23)}} 
                                & PSNR $\uparrow$            & 19.10                      & 17.84                      & 16.92                      & 20.42                        & 17.60                        & 19.03                        & 18.64                                           \\
                                 & SSIM $\uparrow$            & 0.636                      & 0.630                    & 0.616                     & 0.690                          & 0.708                        & 0.664                        & 0.657                                           \\
                                 & LPIPS $\downarrow$         & 0.560                      & 0.520                     & 0.590                      & 0.542                        & 0.566                        & 0.591                        & 0.561                                          \\

          \hline
          \multirow{3}{*}{\makecell{Point-SLAM \cite{pointslam}\\(ICCV23)}} 
                                & PSNR $\uparrow$            & \ud{24.09}               & \ud{22.14}                   & \ud{21.33}                 & 22.74                        & \ud{22.29}             & \ud{24.36}                   & \ud{22.83}                                           \\
                                 & SSIM $\uparrow$                 & 0.715                 & 0.683                      & 0.619                     & 0.630                        & 0.746                        & 0.717                        & 0.685                                         \\
                                 & LPIPS $\downarrow$              & 0.471                 & 0.480                       & 0.554                     & 0.560                        & 0.520                        & 0.512                        & 0.516                                           \\

          \hdashline
          
          \multirow{3}{*}{\makecell{MonoGS \cite{monogs}\\(CVPR24)}} 
                                & PSNR $\uparrow$               & 17.20                      & 15.44                      & 16.92                      & 18.79                        & 13.76                        & 18.10                        & 16.72                                           \\
                                 & SSIM $\uparrow$             & 0.636                      & 0.574                      & 0.657                      & 0.704                        & 0.594                        & 0.705                        & 0.645                                       \\
                                 & LPIPS $\downarrow$          & 0.560                      & 0.628                      & 0.571                      & 0.553                        & 0.723                        & 0.542                        & 0.596                         \\
          \hline
          \multirow{3}{*}{\makecell{SplaTAM \cite{splatam}\\(CVPR24)}} 
                                & PSNR $\uparrow$               & 19.70                      & 19.65                      & 19.11                      & \ud{23.23}                        & 18.58                        & 20.64                        & 20.15                                           \\
                                 & SSIM $\uparrow$             & 0.654                      & \ud{0.807}                      & \ud{0.747}                      & 0.787                        & 0.749                        & 0.749                        & 0.749                                       \\
                                 & LPIPS $\downarrow$          & \ud{0.422}                      & \textbf{0.240}                      & \ud{0.315}                      & \textbf{0.276}                        & \ud{0.351}                        & \textbf{0.278}                        & \ud{0.313}                         \\
          \hline
          \multirow{3}{*}{\makecell{RTG-SLAM \cite{rtg}\\(SIGGRAPH24)}} 
                                & PSNR $\uparrow$               & 18.98                      & 17.44                      & 16.87                      & 19.39                        & 17.51                        & 18.97                        & 18.19                                           \\
                                 & SSIM $\uparrow$             & 0.804                      & 0.705                      & 0.684                      & 0.778                        & 0.716                        & 0.757                        & 0.741                                       \\
                                 & LPIPS $\downarrow$          & 0.488                      & 0.534                      & 0.579                      & 0.499                        & 0.613                        & 0.538                        & 0.542                         \\
          \hline
          \multirow{3}{*}{Ours} 
                                & PSNR $\uparrow$             & \textbf{25.87}               & \textbf{24.04}                      & \textbf{24.86}                      & \textbf{26.84}                        & \textbf{25.91}                        & \textbf{26.13}                        & \textbf{25.61}                                           \\
                                 & SSIM $\uparrow$             & \ud{0.807}               & \textbf{0.836}                      & \textbf{0.864}                      & \textbf{0.838}                        & \textbf{0.873}                      & \ud{0.819}                        & \textbf{0.840}                                                \\
                                 & LPIPS $\downarrow$          & \textbf{0.345}               & \ud{0.270}                      & \textbf{0.271}                      & \ud{0.285}                        & \textbf{0.322}                                        & \ud{0.340}                        & \textbf{0.306}                         \\
          \hline
          \end{tabular}
  }
  \vspace{-5pt}
  \caption{Quantitative comparison of our method against baselines for rendering results on the ScanNet \cite{scannet} dataset.}
  \label{tab:scannet_per}
  \vspace{-5pt}
\end{table}

\begin{figure*}[th]
  \begin{center}
  \includegraphics[width=0.92\linewidth]{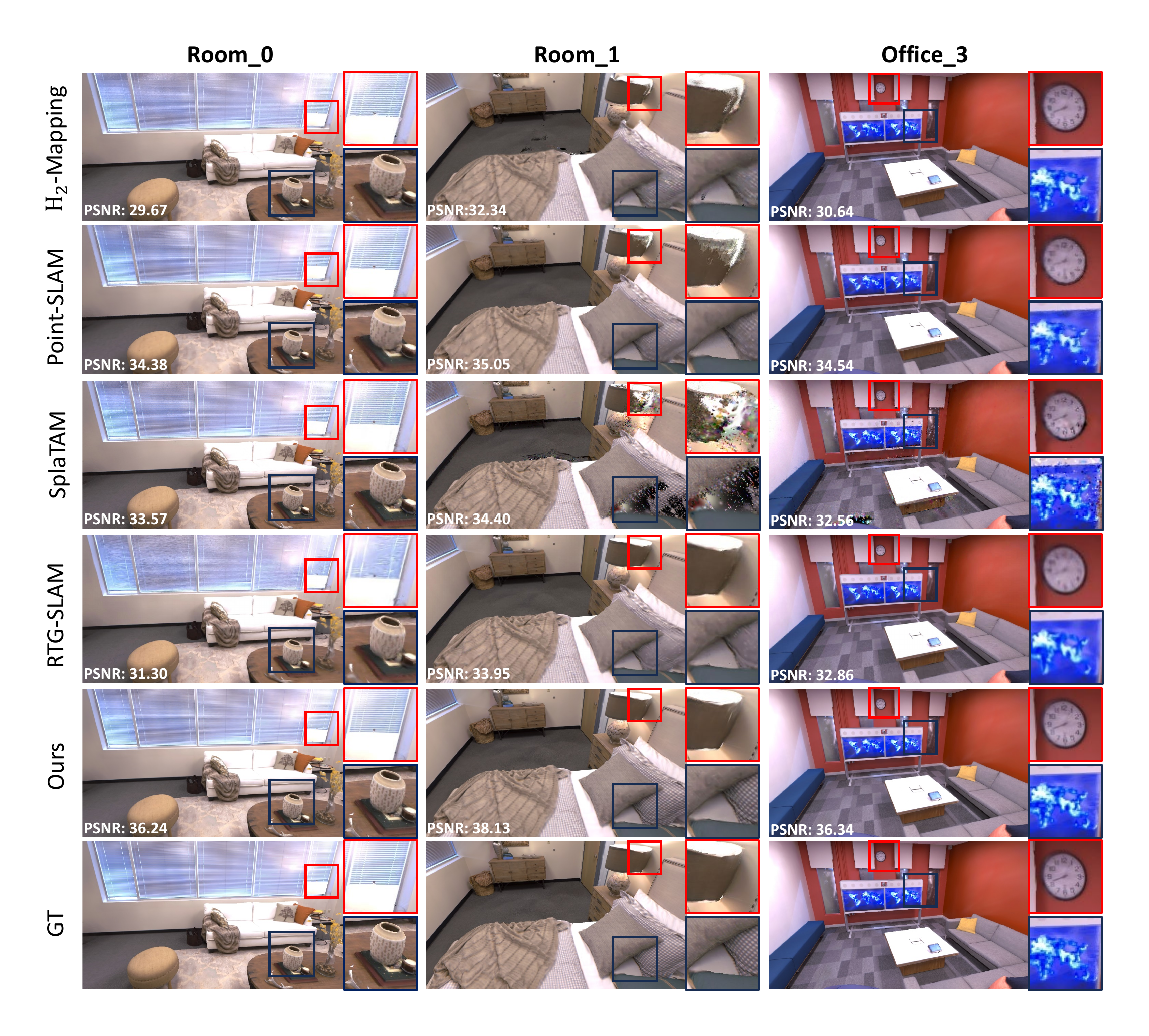}
  \end{center}
  \vspace{-20pt}
  \caption{ {Qualitative comparison of rendering results across three scenes from the Replica dataset. Key details are highlighted by colored boxes.
     The average PSNR metric for each scene is indicated in the lower-left corner.}}
  \vspace{-10pt}
  \label{fig:replica}
\end{figure*}

\begin{table}[h]
  \resizebox{\linewidth}{!}{
      \begin{tabular}{c|ccccccc}
          \hline
           \textbf{Methods}   &  PSNR$\uparrow$                     & SSIM$\uparrow$                & LPIPS$\downarrow$                & \makecell{Depth \\L1(cm)$\downarrow$}    & \makecell{Rendering \\FPS$\uparrow$}                    & \makecell{Model \\Size(MB)$\downarrow
           $} \\
          \hline
                                                      
                      
          {\makecell{MonoGS \cite{monogs}}} 
                                & 25.31            & 0.808                      & 0.543                      & 3.94                      & \textbf{862.0}                        & \ud{5.6}                                                                                   \\

          \hline
          {\makecell{SplaTAM \cite{splatam}}} 
                                & 22.99               & 0.723                      & \ud{0.378}                      & 1.60                      & 54.9                        & 540.3                                                                   \\
          \hline
          {\makecell{RTG-SLAM \cite{rtg}}} 
                                & 24.40               & 0.823                      & 0.415                       & 2.72                      & 316.5                        & 270.8                          \\
          \hline
          {Ours-sparse} 
                                & \ud{28.91}             & 0.846               & 0.423                      & \ud{1.10}                      & \ud{647.8}                        & \textbf{5.4}                                                                   \\                      
          \hline
          {Ours} 
                                & \textbf{30.32}             & \textbf{0.874}               & \textbf{0.300}                      & \textbf{0.89}                      & 559.7                        & 30.5                                                                   \\
          \hline
          \end{tabular}
  }
  \vspace{-10pt}
  \caption{Quantitative comparison in terms of rendering, Depth L1, and memory performance on the real scanned scene provided by  \cite{rtg}.}
  \label{tab:hotel}
  \vspace{-15pt}
\end{table}

\vspace{-5pt}
\subsection{Experiments Results.}{
\noindent
\textbf{Evaluation on Replica \cite{replica}.}{
In Tab. \ref{tab:replica_per}, we evaluate our method's rendering quality results on Replica dataset \cite{replica}. 
Our approach achieves the best PSNR and LPIPS results, while maintaining a highly competitive SSIM result.
Fig. \ref{fig:replica} provides a qualitative comparison of the rendering of ours and baseline method. 
}
Thanks to our hierarchical representation optimization method, our approach is able to better capture details.
Tab. \ref{tab:replica_statis} provides the performance analysis of MonoGS \cite{monogs}, SplaTAM \cite{splatam}, RTG-SLAM \cite{rtg} and our method. 
Since we don't need to explicitly store all 3D Gaussian parameters, our method is more storage-efficient. Despite using fewer iterations and a more sparse distribution of 3D Gaussians, our approach still achieves superior mapping results.

\noindent
\textbf{Evaluation on Real-world Scene.}{
Since the camera trajectory provided by the ScanNet \cite{scannet} originates from BundleFusion \cite{bf} rather than ground truth, 
and the depth images contain significant noise, conducting online mapping on ScanNet is extremely challenging. 
Tab. \ref{tab:scannet_per} presents the rendering evaluation results on six scenes of ScanNet. 
Due to the lack of structured organization, previous Gaussian-based methods \cite{monogs,splatam, rtg} are highly sensitive to depth noise and camera trajectory inaccuracies. In contrast, our method effectively filters out these noises, resulting in more realistic rendering outcomes.
In Tab. \ref{tab:scannet_statis}, we compare the performance of OG-Mapping with other Gaussian-based methods. The results demonstrate that our method can construct more compact maps at a faster processing speed.
Fig. \ref{fig:teaser} and Fig. \ref{fig:scannet} also illustrate that our method can achieve better visual quality with more reliable geometry and texture details. The rendering viewpoints of these results are selected outside of the training views, with random perturbations added to the poses.
Since the depth information provided by ScanNet \cite{scannet} contains excessive noise, the Depth-L1 method used in SplaTAM \cite{splatam} is not suitable for measuring geometric accuracy.
In Tab. \ref{tab:hotel}, we provide additional experimental results on the high-quality large indoor scene  \cite{rtg} to further validate the effectiveness of our method in real-world scenarios. 
}

\begin{table}[h]
  \resizebox{\linewidth}{!}{
      \begin{tabular}{ccccccccc}
          \hline
          \textbf{Methods}                              & PSNR $\uparrow$  & \makecell{Depth L1\\ \space [cm] $\downarrow$}        & \makecell{Processing \\FPS$\uparrow$}           & \makecell{Model \\Size[MB]$\downarrow$}             \\
          \hline
          \multirow{1}{*}{MonoGS \cite{monogs}}          & 35.68                 & 2.12                 & 1.2                    & 9.7                                \\
          \hline
          \multirow{1}{*}{SplaTAM \cite{splatam}}        & 35.78                 &  0.61              &  0.4       & 275.1                              \\
          \hline
          \multirow{1}{*}{RTG-SLAM \cite{rtg}}           & 35.12                 &  1.69              &  15.6                  & 71.5                             \\
          \hline
          \multirow{1}{*}{Ours-sparse\faRunning}                & 36.75                 &  0.72              &  \textbf{16.1}                  & 8.9                 \\ 
          
          \hline
          \multirow{1}{*}{Ours-sparse}                  & 37.00                 &  0.69              &  8.5                  & \textbf{8.8}                             \\
          \hline
          \multirow{1}{*}{Ours}                         & \textbf{38.56}        &  \textbf{0.46}     &  5.6       & 34.6                             \\
          \hline
          \end{tabular}
  }
  
  \vspace{-5pt}
  \caption{Comparison of key performance metrics on the Replica  \cite{replica} dataset between ours and baseline Gaussian-based methods \cite{monogs, splatam, rtg}. }
  \label{tab:replica_statis}
  \vspace{-10pt}
\end{table}
\begin{table}[h]
  \resizebox{\linewidth}{!}{
      \begin{tabular}{cccc}
          \hline
          \textbf{Methods}                              & PSNR $\uparrow$   & \makecell{Processing \\FPS$\uparrow$}    &  \makecell{Model \\Size[MB]$\downarrow$}                 \\
          \hline
          \hline
          \multirow{1}{*}{MonoGS \cite{monogs}}        & 16.72                     & 2.9                          & 5.6                           \\

          \multirow{1}{*}{SplaTAM \cite{splatam}}        & 19.99                     & 1.8                          & 160.7                           \\
          
          \multirow{1}{*}{RTG-SLAM \cite{rtg}}        & 18.19                     & 11.4                          &   153.2                      \\
          
          \multirow{1}{*}{Ours-sparse}                       & 23.87                     & \textbf{12.5}                             & \textbf{3.1}                   \\
          
          \multirow{1}{*}{Ours}                         & \textbf{25.61}                     & 10.1                         & 36.1                            \\
          \hline
          \end{tabular}
  }
  
  \vspace{-8pt}
  \caption{Comparison of key performance metrics on ScanNet \cite{scannet} between ours and baseline Gaussian-based methods \cite{splatam, monogs, rtg}.}
  \label{tab:scannet_statis}
  \vspace{-10pt}
\end{table}
\begin{table}[h]
  \resizebox{\linewidth}{!}{
      \begin{tabular}{ccccc}
          \hline
          \textbf{Methods}                                  & PSNR$\uparrow$                 &  \makecell{Processing\\FPS$\uparrow$}   & \makecell{Model \\Size(MB)$\downarrow$}            \\
          \hline
          \multirow{1}{*}{Projection-based \cite{splatam}}        & 36.18                                             &  2.95    & 71.8                       \\
          \multirow{1}{*}{Naive Unique}                     & 36.24                                            &  3.37    & 39.9                        \\
          \multirow{1}{*}{Octree-based(ours)}               & 36.24                                             &  5.16    & 39.9                         \\
          \hline
          \end{tabular}
  }
  
  \vspace{-8pt}
  \caption{Ablation study of map densification methods quantitative results on room0  \cite{replica}. }
  \label{tab:anchor_manage}
  \vspace{-15pt}
\end{table}

\begin{table}[h]

      \begin{tabular}{c|c|cccccccc}
          \hline
          \textbf{Dataset}                                    & Methods                                   & PSNR$\uparrow$                  & \makecell{Model \\Size(MB)$\downarrow$}   \\
          \hline
          \multirow{3}{*}{Replica}                            & w/o Growing                               & 36.85                           & 29.4                        \\
                                                              & Error-based \cite{scaffoldgs}             & 38.54                           & 40.5                        \\
                                                              & Anchor-based(ours)                        & 38.56                           & 34.6                        \\         
          \hline
          \multirow{3}{*}{ScanNet}                            & w/o Growing                               & 24.89                           & 19.9                        \\
                                                              & Error-based \cite{scaffoldgs}             & 25.27                           & 78.4                        \\
                                                              & Anchor-based(ours)                        & 25.61                           & 36.1                        \\
          
          \hline
          \end{tabular}
  
  \vspace{-5pt}
  \caption{Ablation study of our progressive map refinement strategy on Replica \cite{replica} dataset and ScanNet \cite{scannet} dataset. }
  \label{tab:growing_ablation}
  \vspace{-5pt}
\end{table}

\begin{table}[h]
  \resizebox{\linewidth}{!}{
  \begin{tabular}{c|c|ccccc}
      \toprule
      Type&\multicolumn{1}{c}{Method}  & \multirow{1}{*}{PSNR$\uparrow$} & \multirow{1}{*}{SSIM$\uparrow$} & \multirow{1}{*}{LPIPS$\downarrow$}\\
  
      \hline
      \multirow{3}{*}{\makecell{Fixed \\Window}} & Random \cite{eslam}             & 37.14   &   0.969  &  0.057 \\
      \cline{2-5}
      & Overlap \cite{splatam,monogs}     & 35.85   &   0.963  &  0.063 \\
      \cline{2-5}
      & \makecell{Coverage-\\Maximizing \cite{h2mapping}}     & 37.75   &   0.972  &  0.050 \\
      \hline
      \makecell{Dynamic \\Window}& Global(ours)             & \textbf{38.56}   &   \textbf{0.976}  &  \textbf{0.048}\\
      \bottomrule
  \end{tabular}
  }
  \vspace{-5pt}
  \caption{Ablation study of our dynamic keyframe window methods on Replica \cite{replica} dataset.}
  \vspace{0pt}
  \label{tab:keyframe_performance_comparison}
\end{table}
\begin{figure}[thp]
  \begin{center}
  \includegraphics[width=1.0\linewidth]{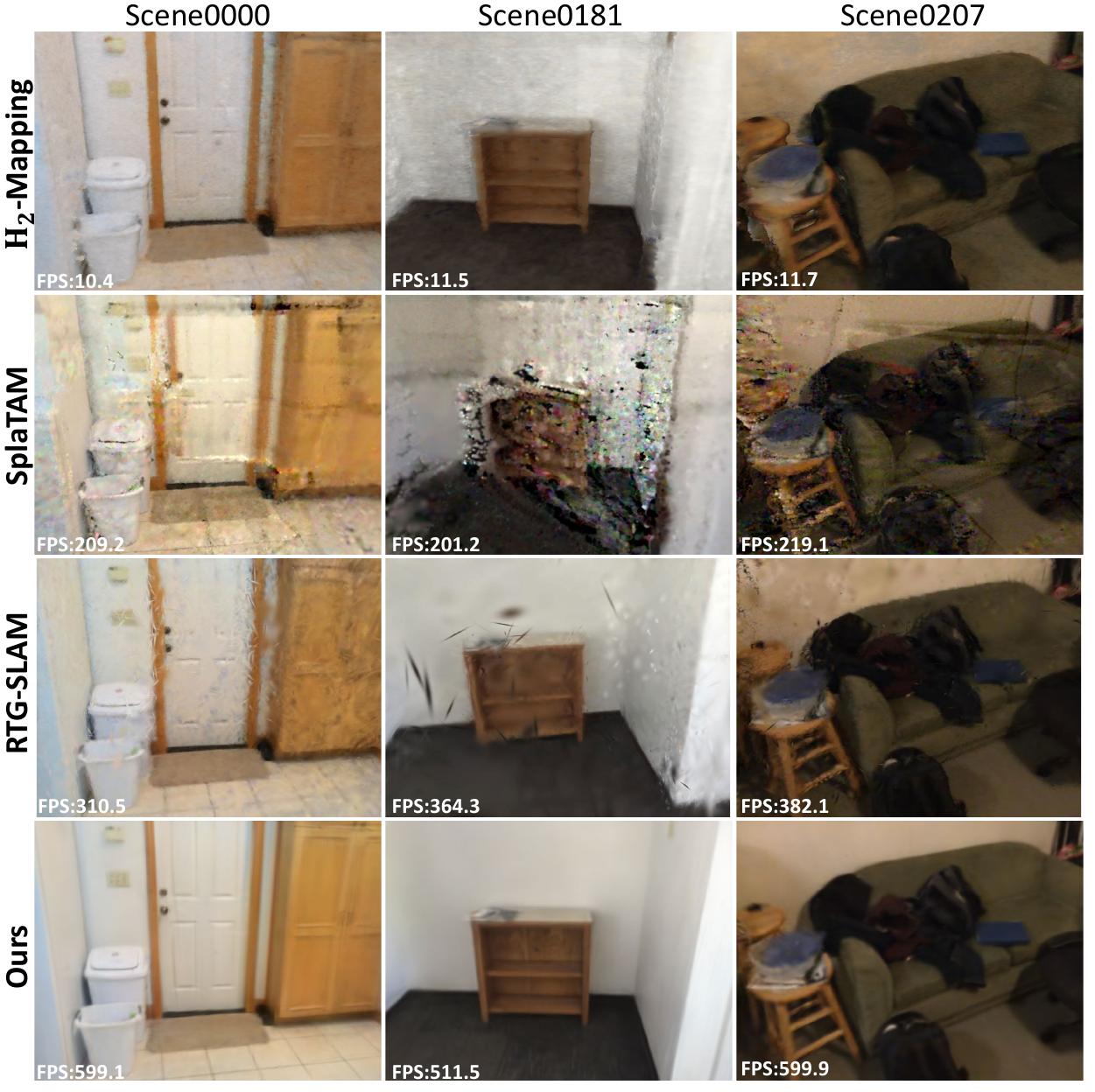}
  \end{center}
  \vspace{-12pt}
     \caption{ {Qualitative comparison of rendering results across three scenes from Scannet. The average rendering FPS are indicated at the bottom of the images.}}
  \vspace{-13pt}
  \label{fig:scannet}
\end{figure}
\noindent
\textbf{Map Densification Ablations.}{
To prove the effectiveness of our octree-based structured map densification method, we compare the mapping performance of Depth error-based method \cite{splatam}, Naive Unique method (Using the unique function to determine whether the anchors are newly observed), and our octree-based method on the room0 of Replica.
As shown in Tab. \ref{tab:anchor_manage},
The projection-based method inevitably adds extra anchors at object edges incorrectly, resulting in larger map size and worse rendering results.
While the Naive Unique method can handle anchor growth correctly like ours, the inefficient unique computation results in an intolerable time overhead.
Our octree-based method leverages the efficiency of sparse octrees, combining both processing accuracy and high efficiency.
}

\noindent
\textbf{Progressive Map Refinement Ablations.}{
  We evaluated our progressive map refinement described in Sec. \ref{sec: optimization}.
  Tab. \ref{tab:growing_ablation} shows the results of disabling growing operation and employing the error-based method  \cite{scaffoldgs} on the Replica and the ScanNet datasets.
  The results show the growing operation is crucial for accurately reconstructing details.  
  Compared to the error-based method \cite{scaffoldgs}, our anchor-based method makes better use of structured information, 
  resulting in smaller map occupancy while maintaining comparable rendering accuracy.
}

\noindent
\textbf{Keyframe Window Ablations.}{
  In Tab. \ref{tab:keyframe_performance_comparison}, we compared our dynamic keyframe window method with other existing keyframe window construction approaches. 
  The overlap method, which only uses keyframes that overlap with the newly added keyframe, is not suitable for addressing the problem of forgetting. 
  Consequently, it results in the poorest rendering quality. The Coverage-Maximizing method  \cite{h2mapping} uses the most recently unoptimized keyframes to construct the keyframe window. However, the reconstruction process tends to overfit these areas as this window is fixed. 
  In contrast, our dynamic keyframe window overcomes these issues, achieving optimal results.

}

}

%% file: sec/5_conclusion.tex
\vspace{-5pt}
\section{Conclusion}
\vspace{-5pt}
\label{sec:conclusion}
In this work, we introduce OG-Mapping, a novel framework for effective online dense mapping.
By utilizing an octree-based structured 3D Gaussians representation, OG-Mapping achieves efficient map densification and compaction. 
Additionally, we propose an anchor-based progressive map refinement strategy to to enhance the capture of finer details.. 
Furthermore, we develop a dynamic keyframe window to mitigate the issues of local overfitting and forgetting problems encountered during the reconstruction process.
Experiment results demonstrate that this approach, leveraging a more compact map, outperforms existing algorithms.
The advantages of our structural representation and dynamic keyframe window are particularly evident in challenging real scenes where existing Gaussian-based online mapping methods typically falter.

\noindent
\textbf{Limitation.}
OG-Mapping relies on inputs from an RGB-D sensor to build a sparse octree. Future research is anticipated to explore the use of monocular image input alone.
Our method employs MLPs as feature decoders, successfully constructing compact maps. 
However, in particularly large scenes, it may still encounter forgetting issues. 
Further research on submap construction \cite{yugay2023gaussianslam,ngm} could further reduce frame processing time.